\documentclass[runningheads]{llncs}
\usepackage[T1]{fontenc}
\usepackage{amsmath, amssymb, booktabs, multirow}
\usepackage{graphicx,verbatim}
\usepackage{xcolor}
\usepackage{subcaption} 
\usepackage{hyperref}

\begin{document}
\title{Bridging MRI and PET physiology:\\ Untangling complementarity through orthogonal representations}

\titlerunning{Bridging MRI and PET}
%

\author{Sonja Adomeit\inst{\ast 1,2} \and
Kartikay Tehlan\inst{\ast 1,2,3,4}\and
Lukas F\"orner\inst{1,2,3,4} \and \\
Katharina Weisser\inst{1} \and
Helen Scholtiseek\inst{5} \and
David Kaufmann\inst{1} \and
Julie Steinestel\inst{6} \and
Constantin Lapa\inst{4,5,7} \and
Thomas Kr\"oncke\inst{1,4,7} \and
Thomas Wendler\inst{1,2,3,4,7}}
\authorrunning{S. Adomeit, K. Tehlan et al.}
\institute{Dept. of diagnostic and interventional Radiology and Neuroradiology, University Hospital Augsburg, Germany \\ \textbf{\email{sonja.adomeit@uk-augsburg.de}}\and
Digital Medicine, University Hospital Augsburg, Germany \and
Chair for Computer Aided Medical Procedures and Augmented Reality, Technical University of Munich, Germany
\and
Bavarian Center for Cancer Research (BZKF) Augsburg, Germany \and
Dept. of Nuclear Medicine, University Hospital Augsburg, Germany \and
Dept. of Urology, University Hospital Augsburg, Germany \and
Center for Advanced Analytics and Predictive Sciences, University of Augsburg, Germany\\
\textsuperscript{$\ast$}Equal contribution}

\maketitle              
\begin{abstract}

Multimodal imaging analysis often relies on joint latent representations, yet these approaches rarely define what information is shared versus modality-specific. Clarifying this distinction is clinically relevant, as it delineates the irreducible contribution of each modality and informs rational acquisition strategies. We propose a subspace decomposition framework that reframes multimodal fusion as a problem of orthogonal subspace separation rather than translation. We decompose Prostate-Specific Membrane Antigen (PSMA) PET uptake into an MRI-explainable physiological envelope and an orthogonal residual reflecting signal components not expressible within the MRI feature manifold. Using multiparametric MRI, we train an intensity-based, non-spatial implicit neural representation (INR) to map MRI feature vectors to PET uptake. We introduce a projection-based regularization using singular value decomposition to penalize residual components lying within the span of the MRI feature manifold. This enforces mathematical orthogonality between tissue-level physiological properties (structure, diffusion, perfusion) and intracellular PSMA expression. Tested on 13 prostate cancer patients, the model demonstrates that residual components spanned by MRI features are absorbed into the learned envelope, while the orthogonal residual is largest in tumour regions. This indicates that PSMA PET contains signal components not recoverable from MRI-derived physiological descriptors. The resulting decomposition provides a structured characterization of modality complementarity grounded in representation geometry rather than image translation.
The code is available under: \url{https://github.com/SonjaA14/inrmri2pet}

\keywords{multiparametric MRI  \and PSMA-PET \and implicit neural representations \and multimodal fusion \and prostate cancer imaging}

\end{abstract}
\section{Introduction}

Multimodal imaging analysis increasingly combines data from distinct modalities, yet the structural relationship between what is shared and what is modality-specific remains insufficiently defined. Existing approaches often optimise cross-modal prediction or joint representations without formalising the boundary between recoverable and irreducible signal components. This boundary has practical implications: it determines the portion of one modality that can be inferred from another and informs rational acquisition strategies.

\begin{figure}
\includegraphics[width=\textwidth]{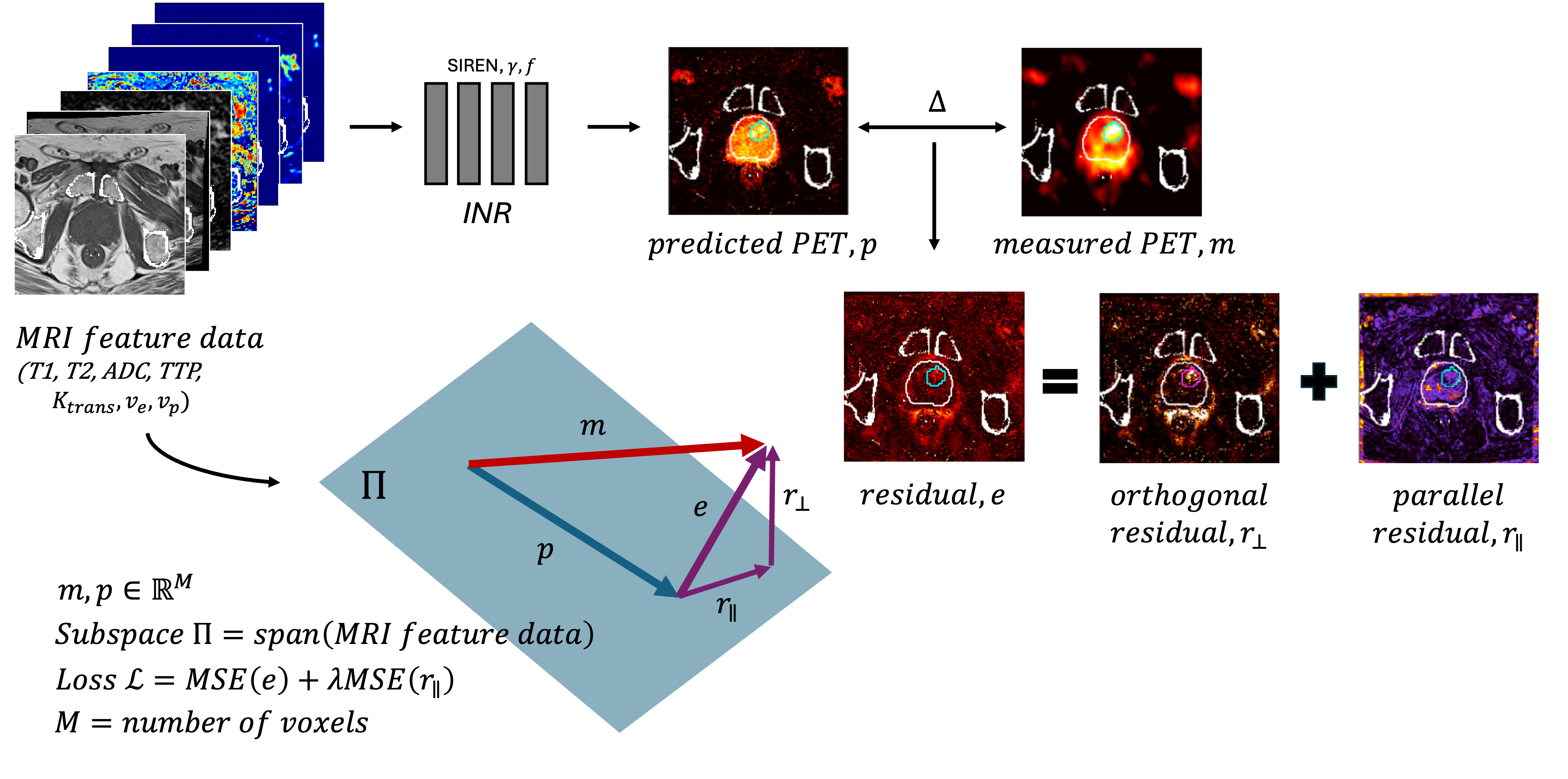}
\caption{Graphical abstract. Seven MRI sequences are used to synthesize a 
PSMA PET image using an INR; synthesis here is a means to 
isolate the orthogonal contribution of PSMA PET with respect to MRI. 
The network is trained with a loss that penalizes the parallel residual 
$r_\parallel$ more heavily than the orthogonal residual $r_\perp$, where 
the total residual $e = r_\parallel + r_\perp$ is decomposed via SVD 
projection onto the subspace $\Pi$ spanned by the MRI feature 
data.}
\label{fig:graphical_abstract}
\end{figure}

We propose a subspace decomposition framework that reframes multimodal fusion as a problem of orthogonal subspace separation. Given a set of source modality features, we decompose the target modality signal into a \textit{physiological envelope}, i.e., a component explainable by the source feature subspace, and an orthogonal residual consistent with processes not directly observable by the source modality (Fig. \ref{fig:graphical_abstract}). Our core contribution lies in a novel loss function that explicitly decomposes the prediction error via Singular Value Decomposition (SVD) into two distinct components: (i) a \textbf{Parallel Residual ($r_{\parallel}$)} that lies within the span of the source feature subspace, representing potentially recoverable but currently unmapped correlations, and (ii) an \textbf{Orthogonal Residual ($r_{\perp}$)}, a component orthogonal to the source feature subspace, representing unique information that cannot be expressed within the span of available source features alone, regardless of model complexity. Moreover, the spatial distribution of $r_{\perp}$ provides a principled, intensity-wise measure of modality complementarity and establishes a theoretical ceiling for cross-modal signal synthesis, defining how much of one modality can ever be recovered from another. 

To parametrise the global mapping in MRI feature space without imposing spatial convolutional priors, we employ an implicit neural representation (INR), implemented as a Sinusoidal Representation Network (SIREN) with Gaussian Fourier Feature encoding~\cite{sitzmann_implicit_2020,tancik_fourier_2020}. 

We validate this framework on the combination of multiparametric MRI (mpMRI) and Prostate-Specific Membrane Antigen (PSMA) PET/CT in prostate cancer. This pair is of particular clinical relevance, as mpMRI characterizes macroscopic tissue architecture, water diffusion, and vascular permeability, while PSMA-PET provides a highly specific molecular signal tied to intracellular protein expression~\cite{maurer_current_2016}. Despite their established diagnostic complementarity, the intensity-level relationship between MRI features and PSMA uptake remains poorly quantified. Standard Uptake Values (SUV) in PET are influenced by a complex interplay of blood flow, interstitial space volume, and receptor density. We hypothesize that these sequences provide a feature subspace describing the macroscopic transport and tissue structure that underlie or modulate PET uptake, while PSMA receptor density constitutes a signal residing outside this physiological landscape.

Using a seven-dimensional MRI feature space ($T_1$, $T_2$, ADC, $K^{\mathrm{trans}}$, $v_e$, $v_p$, voxel-wise Time-to-Peak (TTP) maps), we observe across our cohort that $r_{\perp}$ is significantly higher in tumour regions than in non-tumour tissue. MRI-aligned error is absorbed into the learned envelope, yielding spatially coherent orthogonal residual representation aligned with known tumour biology. 

\section{Related Work}
Multimodal medical imaging has been widely approached through cross-modal synthesis, joint representation learning, and empirical studies of diagnostic complementarity. In cross-modal synthesis, convolutional, adversarial, and more recently diffusion-based models have been used to translate MRI to PET in order to reduce radiation exposure, address missing acquisitions, or generate pseudo-PET references for downstream tasks~\cite{yaakub2019pseudo,chen2026mri,theodorou2025mri2pet}. These approaches demonstrate that substantial PET structure is predictable from MRI features, yet they typically optimise image similarity or task performance rather than defining what fraction of PET signal is fundamentally irrecoverable from MRI. In parallel, multimodal representation learning has focused on separating shared and modality-specific factors in latent space to improve robustness and interpretability~\cite{bousmalis2016domain,lee2021private,wei2024robust}. Domain separation networks and related generative models decompose representations into shared and private components, encouraging invariance while preserving modality-specific content. However, such decompositions are usually implemented in learned latent spaces without an explicit geometric relation to the observed signal fields themselves~\cite{havaei2016hemis,molaei2023implicit}.

Clinically, mpMRI and PSMA PET are recognised as complementary in prostate cancer, with MRI capturing macroscopic tissue architecture, diffusion, and perfusion, and PSMA PET providing molecular information linked to receptor expression and intracellular processes~\cite{tsechelidis2022psma,liu2022prostate}. Combined PET-MRI has been shown to improve diagnostic performance over either modality alone, yet the intensity-wise boundary between MRI-explainable physiology and PET-specific molecular signal remains poorly quantified~\cite{chow2025combined}. INRs have recently been adopted in medical imaging for continuous signal modelling and kinetic parameter estimation, including dynamic PET, highlighting their flexibility in representing high-dimensional physiological functions~\cite{molaei2023implicit,tehlan2025physiological,moussaoui2025implicit}.

\section{Materials and Methods}
We introduce a geometric decomposition defined by the column space of the MRI feature matrix. By projecting the PET prediction residual onto this span via SVD and penalizing the parallel component, we reframe multimodal fusion as orthogonal subspace separation. This yields a structured measure of modality complementarity and defines a formal ceiling on cross-modal recoverability.

\subsection{Intensity-based Learning Formulation}
We define our mapping in MRI feature space rather than anatomical coordinate space, with voxel location serving only to index feature vectors. Let $\mathbf{x}_i \in \mathbb{R}^N$ represent the vector of $N$ MRI-derived features at voxel $i$, and $y_i \in \mathbb{R}$ denote the corresponding normalized SUV. The training set is defined as the collection of pairs $\mathcal{D} = \{(\mathbf{x}_i, y_i)\}_{i=1}^M$, where $M$ is the total number of soft-tissue voxels. This formulation treats each intensity as an independent observation, thereby enabling the network to learn local feature-to-uptake mappings without imposing spatial convolutional inductive bias.

\subsection{Orthogonal Residual Decomposition and Training Objective}
For a feature matrix $\mathbf{X} \in \mathbb{R}^{M \times N}$ and targets $\mathbf{y} \in \mathbb{R}^M$, we decompose the residual $\mathbf{e} = \hat{\mathbf{y}} - \mathbf{y}$ into two orthogonal components 
$\mathbf{e} = \mathbf{r}_{\parallel} + \mathbf{r}_{\perp}$.
In this decomposition, $\mathbf{r}_{\parallel} = \mathbf{P}\mathbf{e}$ represents the projection of the residual onto the column space of $\mathbf{X}$, computed via the projection matrix $\mathbf{P} = \mathbf{X}(\mathbf{X}^\top\mathbf{X} + 10^{-3}\mathbf{I})^{-1}\mathbf{X}^\top$ using SVD, signifying the residual component expressible as a linear combination of the input features. $\mathbf{r}_{\perp} = (\mathbf{I}-\mathbf{P})\mathbf{e}$ is the residual orthogonal to the column space of $\mathbf{X}$. The training objective is
$\mathcal{L} = \text{MSE}(\mathbf{e}) + \lambda \,\text{MSE}(\mathbf{r}_{\parallel})$,
where $\lambda = 1$ empirically.

\subsection{Data Acquisition and preprocessing}
To evaluate the efficacy of this geometric decomposition, we test our approach on 13 patients with histopathologically confirmed 
prostate carcinoma, collected within a retrospective cohort study with local 
ethics committee approval. All subjects underwent mpMRI
and PSMA-targeted PET/CT, with a mean interval of $108.4 \pm 53.2$ days 
between sessions.
The imaging protocol yielded T1w, T2w, ADC, and DCE sequences, from which 
TTP maps and Tofts pharmacokinetic parameters 
were derived. 

TTP is defined voxel-wise as $t_{\hat{k}}$ where $\hat{k} = \arg\max_{k} 
S(\mathbf{x}, t_k)$, with voxels below the 20\textsuperscript{th} percentile 
of peak signal masked to suppress background noise.

Pharmacokinetic parameter maps were computed from DCE sequences using a 
one-compartment Tofts model~\cite{tehlan2025physiological}:
\begin{equation}
C_t(\mathbf{x}, t) = v_p C_p(t) + K^{\mathrm{trans}} \int_0^t C_p(\tau) 
\exp\!\left(-\frac{K^{\mathrm{trans}}}{v_e}(t - \tau)\right) d\tau,
\end{equation}
where $C_p(t)$ is the arterial input function (AIF), $K^{\mathrm{trans}}$ is 
the volume transfer constant, $v_e$ the extravascular extracellular volume 
fraction, and $v_p$ the plasma volume fraction. The baseline signal 
$S(\mathbf{x}, t_0)$ serves as the integration constant, 
and the AIF was derived from iliac arteries segmented via 
TotalSegmentator~\cite{wasserthal_totalsegmentator_2023}.

Lesion volumes (pre-segmented using AutoPET3~\cite{rokuss_fdg_2024}) were corrected by an 
expert, and prostate volumes obtained using 
TotalSegmentator~\cite{wasserthal_totalsegmentator_2023}. The co-registered 
PET SUV volume, normalised by injected dose, physical decay, and body weight, 
serves as the prediction target.

\subsection{Registration and Normalization}

MRI sequences are spatially aligned to the CT reference using a multi-stage pipeline: per-bone rigid registration via signed distance transform optimization, followed by Laplacian interpolation of per-bone transforms to produce a smooth dense displacement field, and finally B-spline deformable registration optimized via mutual information using Elastix to capture residual local deformations. The resulting field is applied to all MRI volumes using linear interpolation.

A valid data mask is defined as the intersection of non-zero voxels across all registered modalities, and volumes are cropped to their bounding box. A soft-tissue mask is generated by thresholding CT to the HU range $[-300,300]$, excluding liquids and cortical bone.

Input features are normalized as follows: T1, T2, ADC, and SUV are 
per-patient min-max normalized to $[0,1]$; TTP maps are divided by 240 
(the maximum acquisition duration in seconds); and Tofts pharmacokinetic 
parameters are Z-score normalized and 
scaled by $1/20$. NaN and Inf values in perfusion maps are replaced with 
zero prior to normalization.


\subsection{Network Architecture}
To mitigate spectral bias and capture high-frequency interactions, we project the input $\mathbf{x}_i$ into a high-dimensional Fourier space using a fixed Gaussian mapping $\phi(\mathbf{x}_i) = [\sin(2\pi \mathbf{B}\mathbf{x}_i), \cos(2\pi \mathbf{B}\mathbf{x}_i)]$, thereby enhancing the network's representation capacity for fine-grained features~\cite{tancik_fourier_2020}.

The mapping $f_\theta: \phi(\mathbf{x}) \to \hat{y}$ is implemented as a Sinusoidal Representation Network (SIREN)~\cite{sitzmann_implicit_2020}, comprising a sinusoidal input layer, three hidden layers (512 units each) of the form $\mathbf{h}_{k+1} = \sin(\omega_0 (\mathbf{W}_k \mathbf{h}_k + \mathbf{b}_k))$ with $\omega_0 = 30$, and a linear output layer. Weights are initialized as proposed in~\cite{sitzmann_implicit_2020}.
Models are optimized over 75 epochs with a batch size of 4096 voxels using the Adam optimizer with AMSGrad ($\eta = 10^{-5}$), and a ReduceLROnPlateau scheduler (factor $= 0.5$, patience $= 5$ epochs) monitoring the training loss. For each patient, two INRs were trained independently: one for Tofts pharmacokinetic parameter estimation and one for PET decomposition. All models were trained on a single integrated 8-core GPU of an Apple M4-based MacBook Air.

\subsection{Evaluation}
Model performance is assessed using Mean Squared Error (MSE) computed separately over three anatomically defined regions: tumour voxels, non-tumoral prostate voxels, and remaining soft-tissue voxels. This regional decomposition allows us to assess whether the orthogonal residual $\mathbf{r}_\perp$ is systematically larger in malignant tissue, as hypothesized.

To evaluate the predictive contribution of individual MRI sequences, we perform systematic ablation studies by selectively withholding input modalities both individually and as semantically grouped combinations: \textit{Structural} (T1w, T2w), \textit{Tofts} ($K^{\mathrm{trans}}, v_e, v_p$), and \textit{Dynamic} (TTP and Tofts maps combined).

\begin{figure}[tbp]
     \centering
     \begin{subfigure}[b]{0.24\textwidth}
         \centering
         \includegraphics[width=\textwidth]{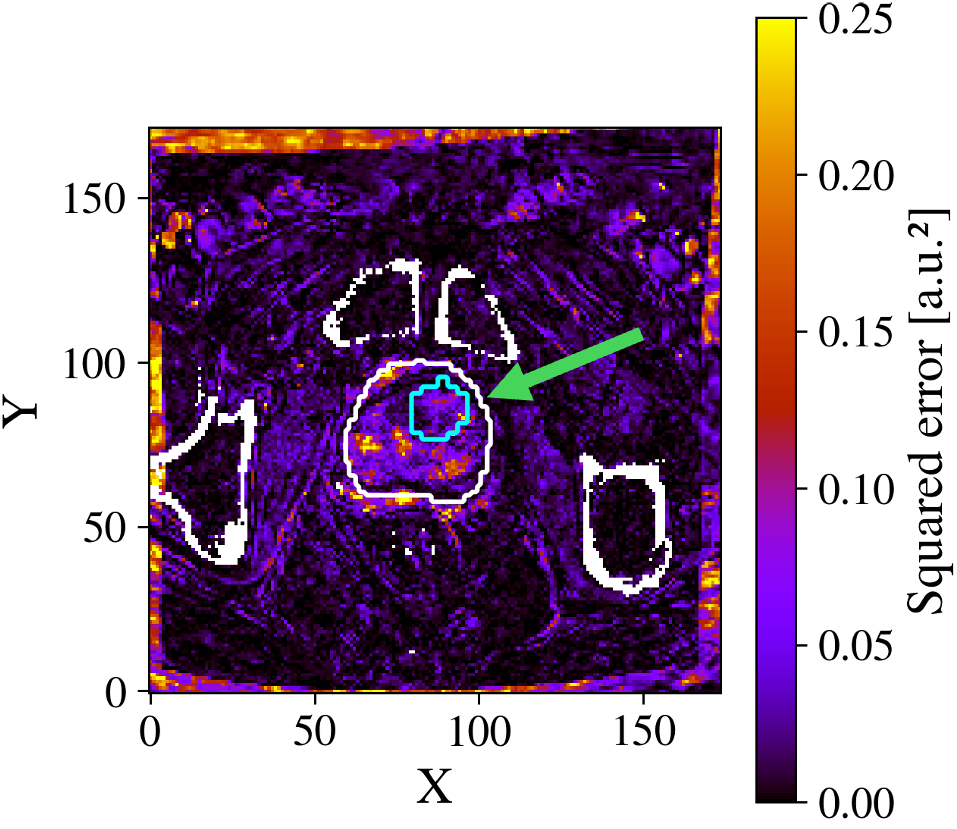}
         \caption{}
         \label{fig:plot_a}
     \end{subfigure}
     \hfill
     \begin{subfigure}[b]{0.24\textwidth}
         \centering
         \includegraphics[width=\textwidth]{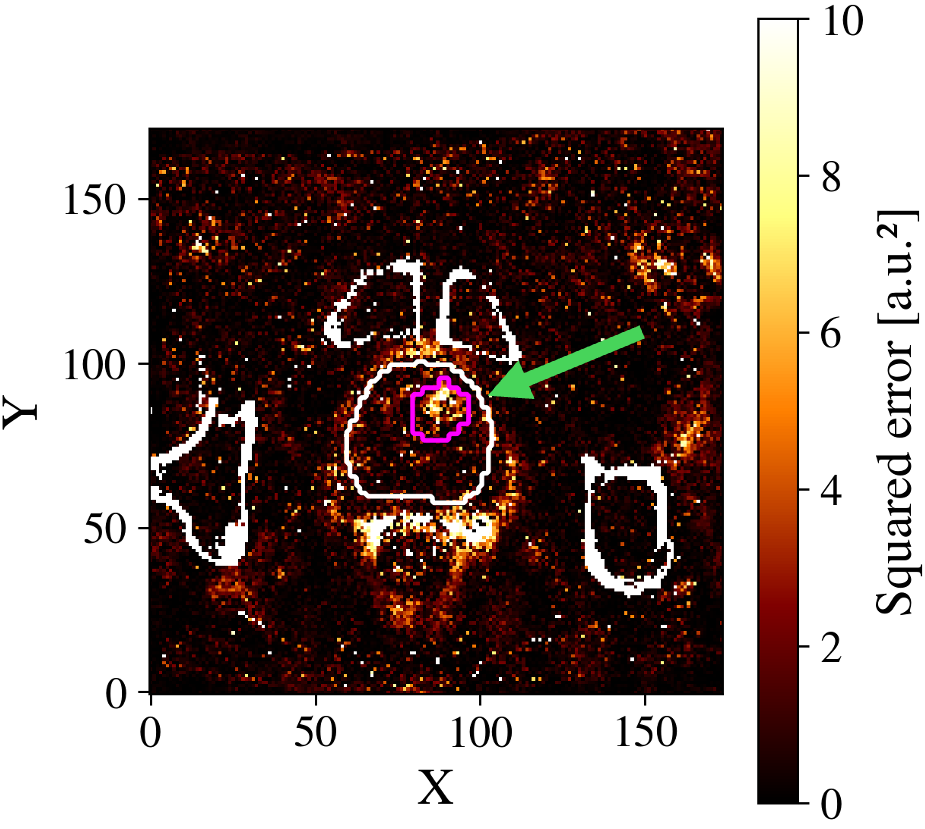}
         \caption{}
         \label{fig:plot_b}
     \end{subfigure}
     \hfill
     \begin{subfigure}[b]{0.24\textwidth}
         \centering
         \includegraphics[width=\textwidth]{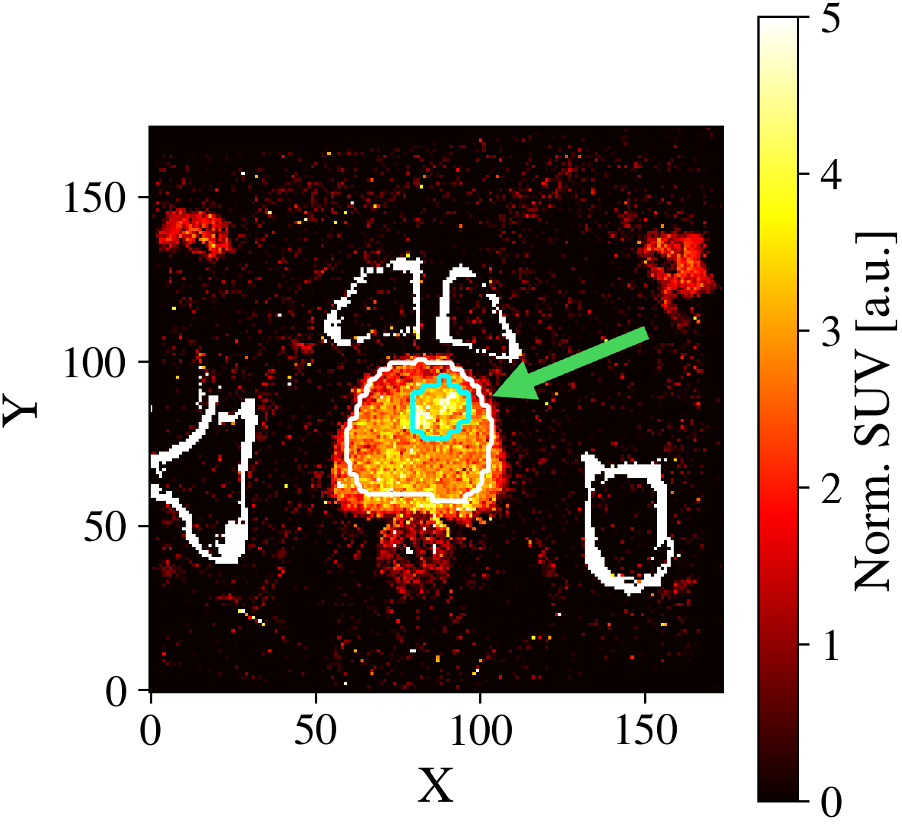}
         \caption{}
         \label{fig:plot_c}
     \end{subfigure}
     \hfill
     \begin{subfigure}[b]{0.24\textwidth}
         \centering
         \includegraphics[width=\textwidth]{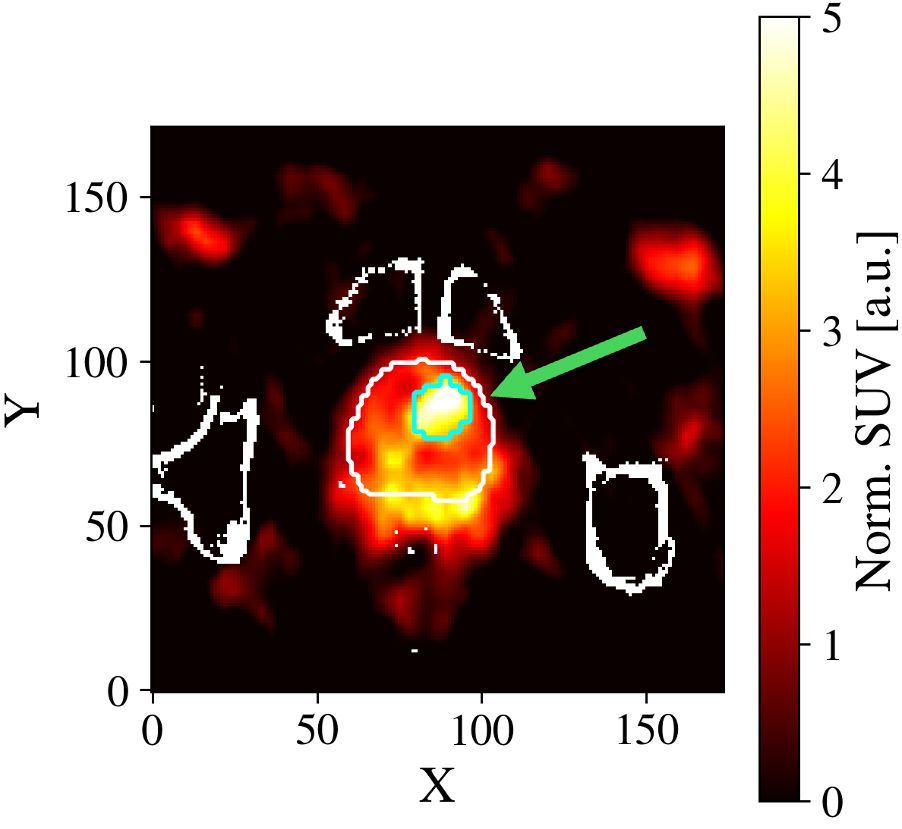}
         \caption{}
         \label{fig:plot_d}
     \end{subfigure}
     
     \caption{Visualization of a patient case featuring: (a) the squared parallel residual; (b) the squared orthogonal residual; (c) the reconstructed SUV slice and (d) the ground truth SUV slice. Prostate masks are delineated white, while tumour ones in cyan (panels (b), (c) and (d)), and in pink (panel (b)).}
     \label{fig2:four_plots}
\end{figure}

\begin{figure}
     \centering
\includegraphics[width=.8\textwidth]{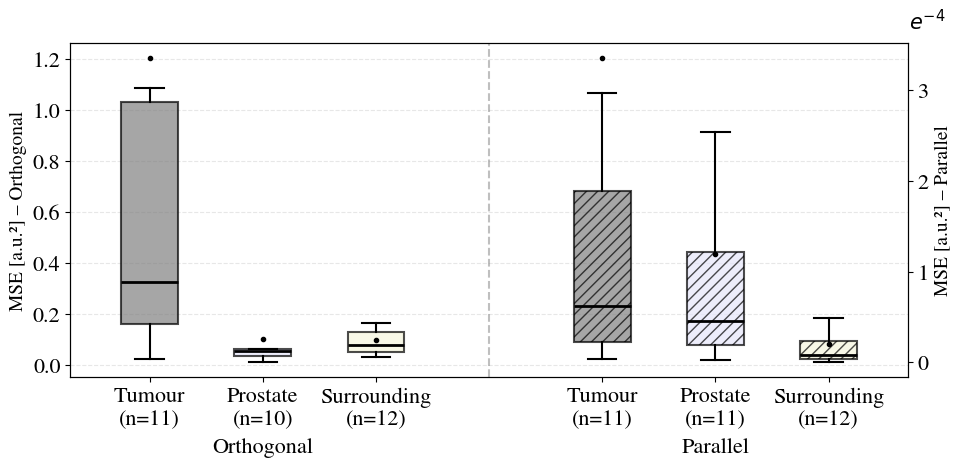}
     \caption{Comparison of orthogonal and parallel error metrics, stratified by tissue type: intra-tumoral regions, non-cancerous prostate tissue, and remaining pelvic anatomy. $n$ state remaining data points after outlier removal for visual clarity.}
     \label{fig3:boxplots}
\end{figure}

\section{Results}

Regional analysis reveals a marked discrepancy in reconstruction fidelity between tissue types. The total MSE in tumour regions ($0.9723$) is approximately $10\times$ higher than in non-tumour regions ($0.0875$ to $0.0942$). Within tumour regions, the orthogonal component ($\text{MSE}_{\perp} = 0.9717$) accounts for $99.9\%$ of the total error, demonstrating that PSMA uptake in malignant tissue is largely orthogonal to the MRI feature space and therefore is an irreducible signal (Fig. \ref{fig3:boxplots} and Tab.\ref{tab:tissue}).

\begin{table}[!t]
\centering
\caption{Quantitative results for the full model for different tissues: differences are statistically significant ($p\ll0.05$).}\label{tab:tissue}
\begin{tabular}{l|c|c|c}
\toprule
\textbf{Tissue} & \textbf{Total MSE$ \pm$SD} & \textbf{MSE$_{\parallel} \pm$ SD} & \textbf{MSE$_{\perp} \pm$ SD} \\
\midrule
Tumour & $0.9723 \pm 6.7172$ & $6.1e^{-4}  \pm 1.4e^{-3}$ & $0.9717  \pm 6.7171$ \\
Prostate & $0.0875 \pm 0.4364$ & $1.3e^{-4}  \pm 4.0e^{-4}$ & $0.0874  \pm 0.4364$ \\
Surrounding & $0.0942 \pm 0.8127$ & $1.6e^{-5}  \pm 1.1e^{-4}$ & $0.0941  \pm 0.8127$ \\
\bottomrule
\end{tabular}
\end{table}

Tab. ~\ref{tab:ablation_results} shows that across all ablation configurations, the orthogonal MSE consistently exceeds the parallel MSE, confirming that the proposed loss formulation successfully drives residual energy into the orthogonal component and away from the MRI-explainable subspace. Individual modality removal yields Total MSE values ranging from $0.0858$ (Minus $v_p$) to $0.1410$ (Minus T2w). Group ablations produce larger performance degradations, with the removal of all dynamic features yielding the highest Total MSE of $0.3028$, highlighting the importance of DCE-derived information for PET envelope prediction. Across all configurations, $\text{MSE}_{\parallel}$ remains consistently low (minimum $7.6 \times 10^{-6}$, achieved without Tofts maps), further validating the effectiveness of the orthogonal decomposition.

\begin{table}[t]
\centering
\caption{Ablation study: Error metrics represent mean $\pm$ standard deviation across the cohort. $\text{MSE}_{\parallel}$ and $\text{MSE}_{\perp}$ denote parallel and orthogonal error components.  Evaluated differences are significant ($p\ll0.05$) between the full model and both leave-one-out (minus $v_p$) and group ablation (no tofts).}
\label{tab:ablation_results}
\begin{tabular}{l|c|c|c}
\hline
\textbf{Ablation Set} & \textbf{Total MSE$ \pm$SD} & \textbf{MSE$_{\parallel} \pm$ SD} & \textbf{MSE$_{\perp} \pm$ SD} \\
\midrule
Full Model & $0.0967 \pm 0.8275$ & $1.9e^{-5} \pm 1.2e^{-4}$ & $0.0967 \pm 0.8275$ \\
\hline
\textit{Leave-one-out} \\
Minus T1 & $0.1323 \pm 0.9724$ & $1.9e^{-5} \pm 1.4e^{-4}$ & $0.1323 \pm 0.9723$ \\
Minus T2 & $0.1410 \pm 0.9674$ & $3.9e^{-5} \pm 3.6e^{-4}$ & $0.1410 \pm 0.9674$ \\
Minus ADC & $0.1369 \pm 1.0131$ & $3.5e^{-5} \pm 4.3e^{-4}$ & $0.1369 \pm 1.0131$ \\
Minus TTP & $0.1228 \pm 0.9017$ & $2.1e^{-5}\pm 1.1e^{-4}$ & $0.1228 \pm 0.9017$ \\
Minus $K^{trans}$ & $0.0907 \pm 0.8090$ & $1.9e^{-5} \pm 2.1e^{-4}$ & $0.0907 \pm 0.8090$ \\
Minus $v_e$ & $0.0920 \pm 0.8449$ & $2.9e^{-5} \pm 2.0e^{-4}$ & $0.0920 \pm 0.8449$ \\
Minus $v_p$ & \boldmath{$0.0858  \pm 0.7193$} & $3.6e^{-5}  \pm 2.8e^{-4}$ & \boldmath{$0.0858  \pm 0.7193$} \\
\hline
\textit{Group Ablation} \\
No Structural & $0.2882 \pm 1.5848$ & $1.3e^{-5} \pm 1.2e^{-4}$ & $0.2882 \pm 1.5848$ \\
No Tofts & $0.1124   \pm 0.9122$ & \boldmath{$7.6e^{-6}  \pm 3.3e^{-5}$} & $0.1124  \pm 0.9122$ \\
No Dynamic & $0.3028 \pm 2.7516$ & $1.0e^{-5} \pm 6.2e^{-5}$ & $0.3028 \pm 2.7516$ \\
\hline
\end{tabular}
\end{table}

\section{Discussion}
This study formalizes multimodal integration as a problem of subspace separation. By defining the column space of MRI-derived features in voxel space and projecting the PET signal accordingly, we decompose uptake into a component expressible within this subspace and an orthogonal remainder. This separation provides a structural distinction between tissue-level physiological variation and PET signal components not representable within the MRI feature space.

The tissue-specific divergence observed in \ref{tab:tissue} quantifies the magnitude of PET signal not representable within the MRI feature subspace. In non-tumour tissue, the Total MSE ranged between $0.0875$ and $0.0942$, whereas in tumour regions it reached $0.9723$, with the orthogonal component accounting for $99.9\%$ of that signal. This structured separation assigns a clear interpretation to the orthogonal component as PET variation not expressible within MRI-derived physiological descriptors. In tumour regions, the approximately ten-fold increase in the orthogonal component, while $\text{MSE}_{\parallel}$ remains minimal ($6.1e^{-4}$), formally separates PET signal components not representable within MRI-derived macro-physiological features, consistent with receptor-mediated processes.
The ablation studies in Tab. \ref{tab:ablation_results} delineate a representational boundary for cross-modal translation. The significant increase in Total MSE when dynamic features were removed (up to a maximum of $0.3028$) suggests that perfusion kinetics and vascular permeability are the primary drivers of the MRI-explainable PET signal. Pure MRI-to-PET synthesis cannot recover molecular processes that lie outside the MRI feature subspace. Our formulation makes this boundary explicit; while structural sequences provide the anatomical framework, they cannot translate into molecular expression.
Rather than learning a monolithic latent space, our framework enforces representational clarity aligned with imaging physics. By using a projection-based regularization, we ensure that the model does not hallucinate PET uptake from MRI noise, but instead separates the signal into distinct physical and biological components. From a theoretical perspective, this formulation connects multimodal imaging with classical projection theory and modern representation learning. It offers a rigorous lens through which to interpret world models in medical AI, where shared and unique information coexist. This framework is highly extensible and can be generalized to other modality pairs, such as CT-to-PET or mpMRI, to systematically map the boundaries of cross-modal information.

Isolating the MRI-aligned envelope may support more targeted use of molecular imaging by identifying regions where PET uptake is not representable within MRI-derived features, to optimise diagnostic and clinical workflows.

\begin{credits}
\subsubsection{\ackname} This research was partially funded by the Intramural Research Funding ``MultiPro'' of the Faculty of Medicine, University of Augsburg, the Bavarian Center for Cancer Research as part of the Lighthouse ``Local Therapies'', as well as by the Bavarian Ministry of Economic Affairs, Regional Development and Energy (StMWi) under grant number DIK-2310-0004// DIK0556/02.

\end{credits}

\newpage
%
%
%
\bibliographystyle{splncs04}
\bibliography{refs.bib}

\end{document}